\documentclass[final]{cvpr}
\pdfoutput=1

\usepackage{times}
\usepackage{epsfig}
\usepackage{graphicx}
\usepackage{amsmath}
\usepackage{amssymb}

\usepackage{bbding}
\usepackage{booktabs}
\usepackage{hyperref}
\hypersetup{
    colorlinks=true,
    linkcolor=red,
    urlcolor=magenta,
}
\usepackage{bbm}
\usepackage{subcaption}

\def\cvprPaperID{****} 
\def\confYear{CVPR 2021}

\begin{document}

\title{SC-Transformer++: Structured Context Transformer for Generic Event Boundary Detection
}

\author{Dexiang Hong$^{1,\ast}$, Xiaoqi Ma$^{1,}$\thanks{Both authors contributed equally to this work.}, Xinyao Wang$^2$, Congcong Li$^{1}$, Yufei Wang$^2$, Longyin Wen$^2$ \\
$^1$University of Chinese Academy of Sciences, Beijing, China.\\
$^2$ByteDance Inc. Mountain View, USA.\\
{\tt\small \{hongdexiang, maxiaoqi.01, xinyao.wang\}@bytedance.com} \\
{\tt\small \{licongcong.lufficc, longyin.wen, yufei.wang\}@bytedance.com}
}

\maketitle

\begin{abstract}
This report presents the algorithm used in the submission of Generic Event Boundary Detection (GEBD) Challenge at CVPR 2022. In this work, we improve the existing Structured Context Transformer (SC-Transformer) method for GEBD. Specifically, a transformer decoder module is added after transformer encoders to extract high quality frame features. The final classification is performed jointly on the results of the original binary classifier and a newly introduced multi-class classifier branch. To enrich motion information, optical flow is introduced as a new modality. Finally, model ensemble is used to further boost performance. The proposed method achieves $86.49\%$ F1 score on Kinetics-GEBD test set. which improves $2.86\%$ F1 score compared to the previous SOTA method. Code is available at \href{https://github.com/lufficc/SC-Transformer }{https://github.com/lufficc/SC-Transformer}.
\end{abstract}


\begin{figure*}[t]
    \centering
    \includegraphics[width=\linewidth]{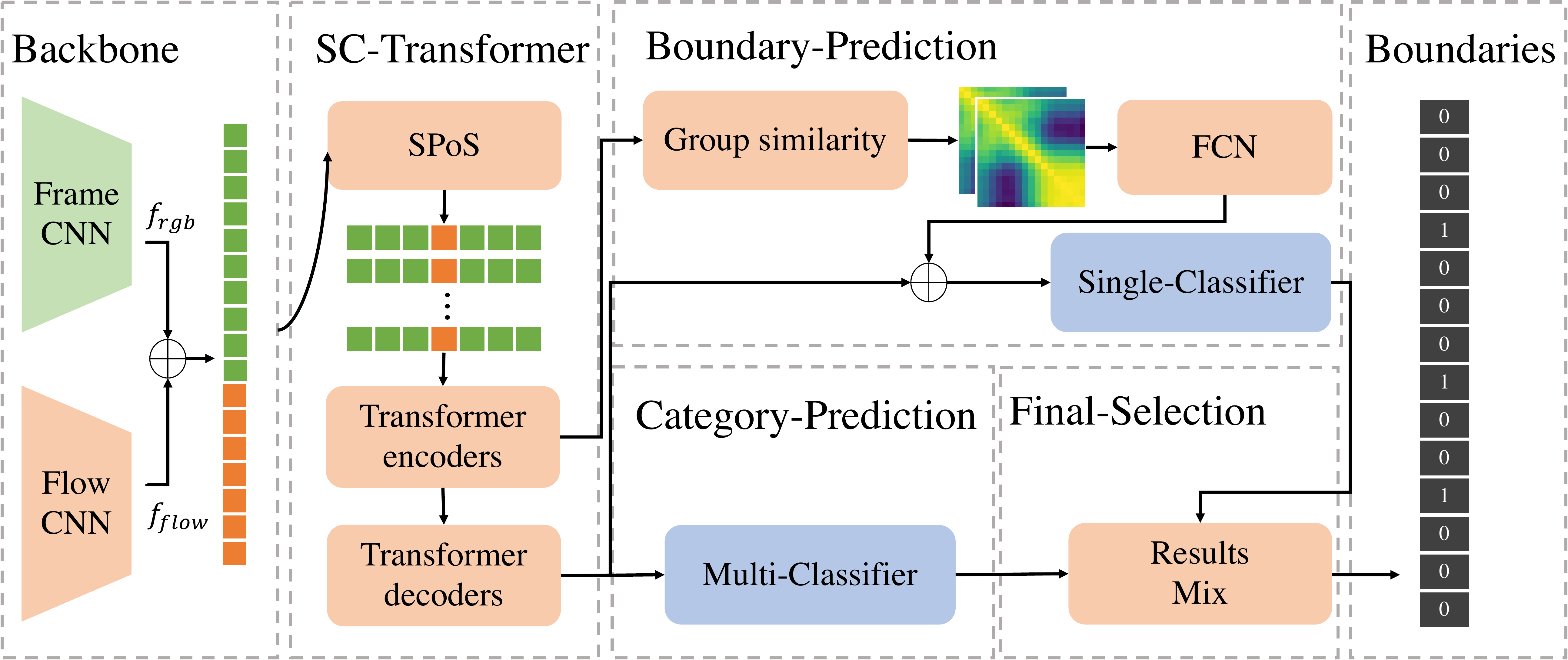}
    \caption{Overview architecture of the proposed method.}
    \label{fig:framework}
\end{figure*}

\section{Approach}
The task of \textbf{Generic Event Boundary Detection (GEBD)} is introduced by \cite{DBLP:journals/corr/GEBD_dataset}, which aims to localize the moments where humans naturally perceive taxonomy-free event boundaries that break a longer event into shorter temporal segments. In the LOVEU@CVPR'21 challange\footnote{\url{https://sites.google.com/view/loveucvpr21}}, we sample frames within a local window and process them individually in multiple forward passes to extract frame features. However, this brings huge computation cost and results in low efficiency training and inference. Besides, since most sampled frames are non-boundary frames, the class imbalance issue must be handled carefully.

The recent proposed Structured Context Transformer (SC-Transformer) \cite{sc-transformer} is an end-to-end method that predicts all sampled video frames in a single forward pass with high efficiency and accuracy. Despite the performance of the SC-Transformer is already the state-of-the-art, things could be done to further boost its accuracy. The overall architecture of our proposed method is presented in Figure \ref{fig:framework}. In our architecture, the structured partition of sequence and group similarity are identical to the original SC-transformer. We recommend readers refer to \cite{sc-transformer} for more detailed information. To summarize, the SC-Transformer is designed for GEBD with its structured partition of sequence (SPoS) mechanism, which has linear computational complexity with respect to input video length and enables feature sharing by design. The SPoS mechanism structures the sliding-window based local feature sequences and each frame to an one-to-one manner, which is termed as \textbf{structured context}. Several modifications are introduced to the original design of the SC-Transformer. First, an optical flow feature extractor is added to work jointly with the RGB feature extractor. Then we modified the SC-Transformer module. More specifically, inside the SC-Transformer module, given a video clip of arbitrary length, we use our modified CSN to extract the temporal and spacial feature representation for each frame and acquire the frame feature sequence, \ie, $V = \{ I_t\}_{t=1}^T$, where $I_t \in \mathbb{R}^C$ and $T$ is the length of the video clip. Then the structured partition of sequence (SPoS) mechanism is employed to re-partition input frame sequence $\{ I_t\}_{t=1}^T$ and provide \textbf{structured context} for each candidate frame. The Transformer \cite{DBLP:conf/nips/att_is_all_you_need} encoder blocks are then used to refine the representation of each partition of the sequence. Different from the original design, the output from the encoders are then applied to newly introduced transformer decoder blocks to extract the overall feature representation of each local sequence. After that, we compute the group similarities to capture temporal changes. Finally, two parallel lightweight fully convolutional networks are used to generate the accurate location and category of each boundary. The classification results from two branches are then merged to make final prediction.

\subsection{Backbone Networks}
The Channel-Separated Convolutional Networks (CSN) \cite{CSN} is first designed for video classification, which separates the channel interactions and spatiotemporal interactions to balance the accuracy and efficiency. To achieve this, all the convolutional operations in CSN are separated into the pointwise $1\times1\times1$ convolutions for channel interactions and depthwise $3\times3\times3$ convolutions for local spatiotemporal interactions. To extract more discriminative features for GEBD, we use CSN as the backbone network to extract features from the RGB domain.

The motion cue in video understanding task is crucial. Inspired by existing methods, we also introduce optical flow information into the framework. The MC3 \cite{MC3} network is used to extract optical flow characteristic information, and then the optical flow features and the RGB features are concatenated as the sampled frame features.

\subsection{SC-Transformer with Decoders}
\label{sec:sc_transformer}
Different types of boundaries usually exhibit inconsistent speed and duration, which further increases the temporal diversities of event boundaries. These spatio-temporal
diversities lead to overly complicated variations in videos, which impedes the accurate detection of event boundaries. Neighbouring feature similarities are crucial cues to identify the locations of event boundaries, however, we argue that they are insufficient to accurately localize all types of event boundaries. Predictions solely based on feature similarities tend to have high precision but low recall on the validation and the test partitions of GEBD. To remedy this, we add transformer decoders to extract the overall feature of the candidate frame from its structured context. After acquiring structured context temporal representation $x_t \in \mathbb{R}^{L \times C}$ from transformer encoders, we use a transformer decoder to capture the overall semantic features of the structured context. The key and value of the transformer decoders are the output from the transformer encoders. We introduce a learnable embedding $q \in \mathbb{R}^{1 \times C}$ as the input query. Then, we can extract the feature embedding $f_t \in \mathbb{R}^{C}$  within the structured context window from the transformer decoder. This feature is then concatenated with the output from the group similarity module to make final prediction.

\begin{table*}[!t]
    \centering
    \begin{tabular}{c c c c c c c}
    \toprule[1.2pt]
        Model & RGB Input & Optical Flow Input & Size 224 & Size 256 & Size 320 & Category Prediction \\
    \hline
        1 & \checkmark & \quad & \checkmark & \quad & \quad & \quad  \\
        2 & \checkmark & \checkmark & \checkmark & \quad & \quad & \quad  \\
        3 & \checkmark & \quad & \quad & \checkmark & \quad & \quad  \\
        4 & \checkmark & \checkmark & \quad & \checkmark & \quad & \quad  \\
        5 & \checkmark & \quad & \quad & \quad & \checkmark & \quad  \\
        6 & \checkmark & \checkmark & \quad & \quad & \checkmark & \quad  \\
        7 & \checkmark & \quad & \checkmark & \quad & \quad & \checkmark  \\
        8 & \checkmark & \checkmark & \checkmark & \quad & \quad & \checkmark  \\
        9 & \checkmark & \quad & \quad & \checkmark & \quad & \checkmark  \\
        10 & \checkmark & \checkmark & \quad & \checkmark & \quad & \checkmark  \\
        11 & \checkmark & \quad & \quad & \quad & \checkmark & \checkmark  \\
        12 & \checkmark & \checkmark & \quad & \quad & \checkmark & \checkmark  \\
        
    \bottomrule[1.2pt]
    \end{tabular}
    \caption{The settings of different ensemble models}
    \label{tab:ensemble}
\end{table*}

\subsection{Boundary Category Classification}
\label{sec:optim}
Although boundaries in GEBD are defined taxonomy-free, the main category of all the boundaries are also provided in the annotation. We believe these categorical labels would also be a useful supervision signal for boundary detection. In addition to the original binary classification head, our model has another multi-class classification branch. Two branches work in parallel, one is responsible for boundary prediction and the other is responsible for boundary category prediction. The Transformer decoder feature $f$ and group similarity feature $h$ are concatenated to do the boundary prediction. Specifically, we use the features from the output of the transformer decoder to predict the category of each boundary, \textit{i.e.}, $b = B(h, f)$, and $m=M(f)$, where $B(\cdot, \cdot$) and $M(\cdot)$ are the binary classification head and multi-class classification head. $b \in \mathbb{R}^{T}$ and $m \in \mathbb{R}^{(C+1) \times T}$ indicate the predictions of boundaries and the corresponding categories. $C$ indicates the number of boundary category. Then we select the max confidence score from these two branch to merge the results. 

\begin{equation}
    p_t = \max(b_t, 1 - m_t[0])
\end{equation}
where $p_t$ indicates the final prediction score.

Similar to \cite{sc-transformer}, we use the Gaussion distribution to smooth the ground-truth boundary labels and obtain the soft labels instead of using the ``hard labels'' of boundaries. Finally, cross entropy loss is used to minimize the difference between model predictions and the soft labels.

\subsection{Post Processing}
During inference, we first obtain the boundary confidence sequence of the whole
video. To prevent duplicate predictions of adjacent frames, we apply peak-estimation technique on the confidence scores. We first select the maximum prediction score within a local window range of $[-4, 4]$. Then a threshold is used to filter out peaks with lower scores, and the peaks with scores higher than the threshold are selected as boundaries.

\subsection{Model Ensemble}
To further boost the performance of our model, we average the prediction scores of several models trained with different architectures and hyperparameters, \eg, we use different modalities of input (RGB and optical flow),  different input size range from 224 to 320 for model training, and whether contain category prediction. The settings of different models are list in Table \ref{tab:ensemble}.

\section{Experiments}

\subsection{Implementation Details}

We use CSN \cite{CSN} pretrained on the IG-65M \cite{IG} and kinetics-400 \cite{kay2017kinetics} as the RGB feature encoder. The pretrained checkpoint is released in the mmaction2 repository\footnote{\url{https://github.com/open-mmlab/mmaction2}}. Different from the original setting, we modify the temporal strides of the CSN backbone from $[1, 2, 2, 2]$ to $[1, 1, 1, 1]$. As a result, the CSN backbone produces feature representations with the same temporal resolution as the input. Images are resized to 256$\times$256. We uniformly sample 100 frames from each video for batching purpose, \ie, $T=100$. We use the standard SGD with momentum set to $0.9$, weight decay set to $10^{-4}$, and learning rate set to $10^{-2}$. We set the batch size to $3$ (3 videos, equivalent to 300 frames) for each GPU and train the model on $8$ NVIDIA Tesla V100 GPUs, resulting in a total batch size of $24$. Automatic mixed precision training is turned on to reduce the memory burden. The network is trained for $20$ epochs with a learning rate drop by a factor of $10$ after $6$ epochs and $10$ epochs, respectively. 

We randomly sample $2000$ videos from kinetics-GEBD validation set to construct local validation set and use all the rest data for training. We refer this set as the mini-val set in the following sections.

\begin{table}[!t]
    \centering
    \begin{tabular}{c c c c}
    \toprule[1.2pt]
        Model & F1 & Precision & Recall \\
    \hline
        SC-Transformer & 0.8312 & 0.8548 & 0.8089 \\
        + Post Processing & 0.8358 & 0.8210 & 0.8512 \\
        + Transformer Decoder & 0.8373 & 0.8323 & 0.8422 \\
        + Category Prediction & 0.8401 & 0.8321 & 0.8482 \\
        + Ensemble & 0.8541 & 0.8380 & 0.8708 \\
    \bottomrule[1.2pt]
    \end{tabular}
    \caption{The results of ablation studies on our mini-val set. \cite{DBLP:journals/corr/GEBD_dataset}}
    \label{tab:ablation}
\end{table}

\begin{table}[!t]
    \centering
    \begin{tabular}{c c c c}
    \toprule[1.2pt]
        Model & F1 & Precision & Recall \\
    \hline
        Baseline\cite{DBLP:journals/corr/GEBD_dataset}$^{\ast}$ & 0.6250 & 0.6240 & 0.6260 \\
        SC-Transformer & 0.8237 & 0.8990 & 0.7600 \\
        \textbf{Ours} & 0.8649 & 0.8745 & 0.8556 \\
    \bottomrule[1.2pt]
    \end{tabular}
    \caption{The final results on the test set. $^{\ast}$Baseline results evaluated on the validation set.}
    \label{tab:final}
\end{table}

\subsection{Ablation Studies}
In this section, we conduct several ablation experiments to comprehensively understand the performance contribution of each module from our proposed framework. The results are shown in Table \ref{tab:ablation}. We report the results on the mini-val validation set. With these methods, we improve the F1 score from 0.8312 to 0.8541 compared to the original SC-Transformer.

\subsection{Test Set Results}
We ensemble $12$ models with different parameters and submit our results on the Kinetics-GEBD test set and achieve $86.49$ F1 score. The results compared with other methods are shown in Table \ref{tab:final}.



{\small
\bibliographystyle{ieee_fullname}
\bibliography{cvpr}
}

\end{document}